\documentclass[preprint,12pt]{elsarticle}

\usepackage{amssymb}
\usepackage{amsthm}
\usepackage{amsmath}
\usepackage{algorithm}
\usepackage{algpseudocode}

\usepackage{url}

\newcommand*{\img}[1]{%
    \raisebox{-.3\baselineskip}{%
        \includegraphics[
        height=\baselineskip,
        width=\baselineskip,
        keepaspectratio,
        ]{#1}%
    }%
}

\journal{Information Sciences}

\begin{document}

\begin{frontmatter}



\title{Dropout Injection at Test Time for Post Hoc Uncertainty Quantification in Neural Networks}

\cortext[cor1]{Corresponding author.}
\author[inst1,inst2]{Emanuele Ledda\corref{cor1}}
\ead{emanuele.ledda@uniroma1.it}
\author[inst3]{Giorgio Fumera}
\ead{fumera@unica.it}
\author[inst2]{Fabio Roli}
\ead{fabio.roli@unige.it}

\affiliation[inst1]{organization={Department of Computer, Control and Management Engineering, Sapienza University of Rome}, 
            addressline={Via Ariosto 25},
            city={Rome},
            postcode={00185},
            country={Italy}}
\affiliation[inst2]{organization={Department of Informatics, Bioengineering, Robotics, and Systems Engineering, University of Genova}, 
            addressline={Via Dodecaneso 35}, 
            city={Genova},
            postcode={16146},
            country={Italy}}
\affiliation[inst3]{organization={Department of Electric and Electronic Engineering, University of Cagliari}, 
            addressline={Via Marengo 3}, 
            city={Cagliari},
            postcode={09100}, 
            country={Italy}}

\begin{abstract}
    Among Bayesian methods, Monte-Carlo dropout provides principled tools for evaluating the \emph{epistemic} uncertainty of neural networks.
    Its popularity recently led to seminal works that proposed activating the dropout layers only during inference for evaluating uncertainty.
    This approach, which we call dropout \emph{injection}, provides clear benefits over its traditional counterpart (which we call \emph{embedded} dropout) since it allows one to obtain a post hoc uncertainty measure for any existing network previously trained without dropout, avoiding an additional, time-consuming training process.
    Unfortunately, no previous work compared injected and embedded dropout; therefore, we provide the first thorough investigation, focusing on regression problems.
    The main contribution of our work is to provide guidelines on the effective use of injected dropout so that it can be a practical alternative to the current use of embedded dropout. In particular, we show that its effectiveness strongly relies on a suitable scaling of the corresponding uncertainty measure, and we discuss the trade-off between negative log-likelihood and calibration error as a function of the scale factor.
    Experimental results on UCI data sets and crowd counting benchmarks support our claim that dropout injection can effectively behave as a competitive post hoc uncertainty quantification technique.
\end{abstract}




\end{frontmatter}


\section{Introduction}
\label{sec:introduction}

Systems based on Artificial Intelligence are nowadays widespread in industry, workplaces and everyday life.
They often operate in critical contexts where errors can cause considerable harm to humans (e.g., self-driving cars), where transparency and accountability are required (e.g., medical and legal domains), and where end users need to be aware of the uncertainty/reliability of AI predictions (e.g., intelligent video surveillance).
In particular, given that AI moved out of research laboratories to real-world scenarios, the need for trustworthiness guarantees led to the growth of an entire research field devoted to quantifying \emph{uncertainty} of machine learning models~\cite{Hullermeier2021,Gal2016thesis}.

In this field, it is common to distinguish between two possible sources of uncertainty:
one caused by intrinsic data noise (\textit{aleatoric} uncertainty, or data uncertainty) and one caused by the lack of knowledge on the ``correct'' prediction model (\textit{epistemic} uncertainty, or model uncertainty)~\cite{KendallGal2017}.
In particular, most of the work on deep neural networks uses their Bayesian extension for quantifying \emph{uncertainty}, modeling the network weights with a probability distribution~\cite{MacKay1992,Neal1995}.
The simple and effective Monte Carlo Dropout technique ~\cite{GalGahahr16} is one of the leading and most widely used techniques for evaluating \emph{epistemic} uncertainty according to the Bayesian extension of neural networks. It is based on \emph{dropout}, a stochastic regularization technique designed for neural networks, consisting of randomly dropping out some neural units and their associated connections during training, with a predefined
\begin{figure}[tb]
    \centering
    \includegraphics[width=\textwidth]{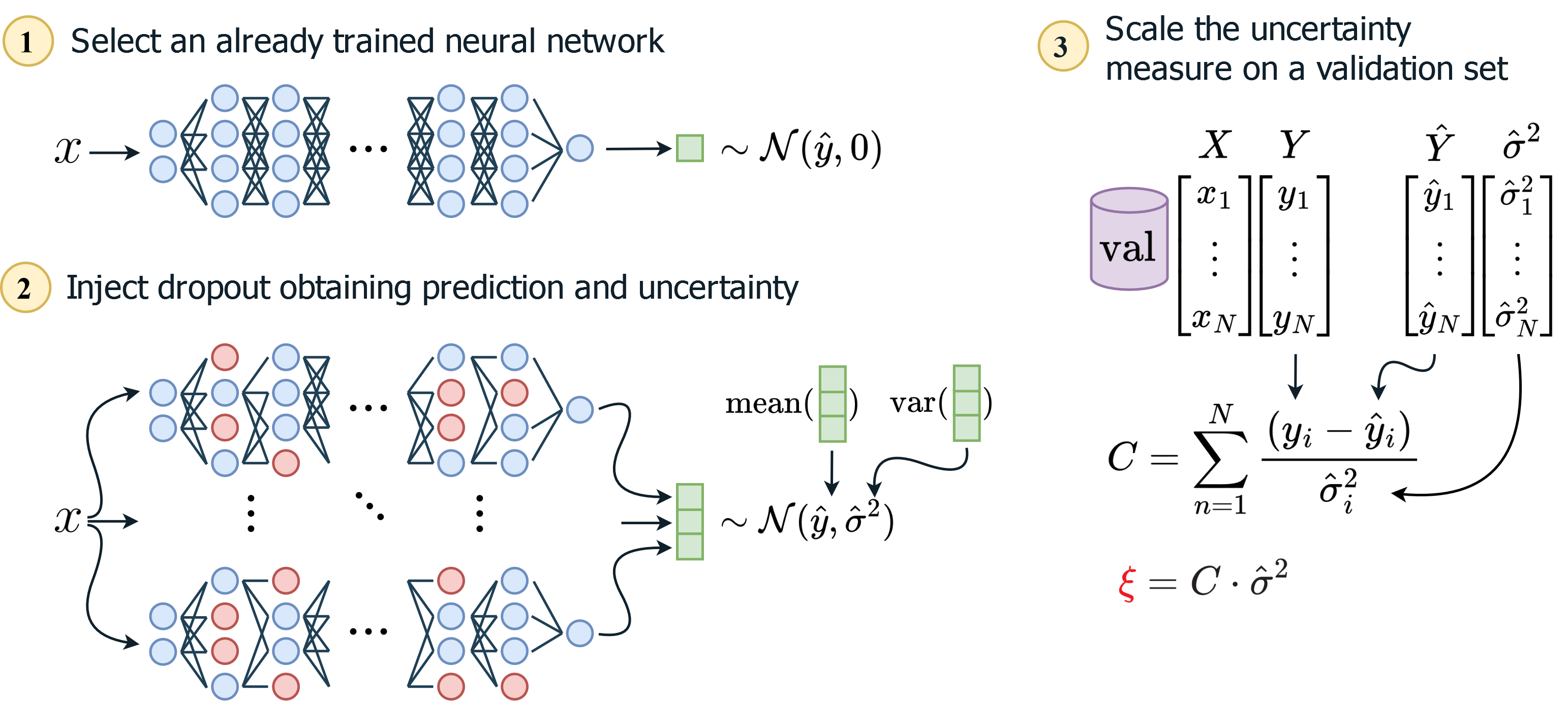}
    \caption{A schematic diagram of the three main steps that we propose to effectively implement injected dropout:
    1) select a neural network trained for a regression problem without dropout (network output: $\hat{y} \in \mathcal{R}$);
    2) add dropout layers to the network;
    3) compute the dropout rate $\Phi$ and the scaling factor $C$ of the uncertainty measure by solving the optimization problem in ~\eqref{eqn:laves} for obtaining the scaled measure $\xi^2$.}
    \label{fig:illustration_injection}
\end{figure}probability~\cite{JMLR:v15:srivastava14a}. 
The main idea behind Monte Carlo Dropout is to exploit the weight randomization induced by the dropout layers by keeping them active during inference; this allows one to approximate the distribution of the neural network weights with respect to training data.
However, whereas dropout has been originally designed as a stochastic regularization technique to be used only during training, when it is used for uncertainty quantification its purpose is to estimate the weight distribution, and it is no more conceived as a regularizer.
This suggests activating dropout layers during training may not be necessary if dropout is used only for uncertainty quantification.
Accordingly, a recent work by Loquercio et al.~\cite{Loquercio2020} proposed to activate dropout layers for uncertainty quantification only during inference, instead of using it also during training. 
The latter approach, which we call dropout \textit{injection}, appears very interesting, as it would allow the \emph{integration} of a measure of epistemic uncertainty into \emph{any} network that was previously trained with no dropout, without the computational burden of a further training procedure required by the original Monte Carlo Dropout, which we call \textit{embedded} dropout, nor the need to optimize the related hyper-parameters.
 
Loquercio et al.~\cite{Loquercio2020} provided empirical evidence that injected dropout, combined with assumed density filtering~\cite{DBLP:conf/uai/BoyenK98} for aleatoric uncertainty evaluation, can be an effective tool for evaluating the \emph{predictive} uncertainty of a predictor.
However, no previous work, including the one by Loquercio et al.~\cite{Loquercio2020}, provided a thorough analysis of injected dropout as an epistemic uncertainty evaluation technique, nor a comparison with classical embedded dropout, in order to provide guidelines for its effective use.

Based on the above motivations, in this work we carry out a deep investigation of injected dropout, focusing on regression problems.
First, our analysis shows that its effectiveness critically relies on a suitable scaling of the corresponding uncertainty measure, differently from embedded dropout.
To this aim, we extend the original formulation by Loquercio et al.~\cite{Loquercio2020} to introduce and compute a suitable scaling factor~\cite{DBLP:conf/midl/LavesIFKO20}.
We also address the resulting issue of achieving a trade-off between different aspects related to the quality of the corresponding uncertainty measure, i.e., negative log-likelihood and calibration error, as a function of the scale factor.
We then carry out experiments on eight UCI data sets as well as five benchmarks for crowd counting, which is a challenging computer vision task, to thoroughly evaluate the behavior of injected dropout in terms of prediction accuracy and the quality of the corresponding uncertainty measure, providing experimental evidence of its effectiveness over embedded dropout.

Our results shed light on the practical operation of injected dropout and show that it can effectively evaluate epistemic uncertainty at test time without a time-consuming training process with dropout layers.

In the following, we present background concepts on Bayesian methods for uncertainty evaluation and Monte Carlo Dropout in Sect~\ref{sec:statement}, a theoretical analysis of injected dropout and the proposed method for injected dropout in Sect.~\ref{sec:dropout_injection}, and
experimental results in Sect.~\ref{sec:experiments}.
Discussion and conclusions are given in Sect.~\ref{sec:conclusions}, where future work on this topic is also discussed.
\section{Background Concepts}
\label{sec:statement}

We first summarise the main concepts of uncertainty quantification in neural networks based on Bayesian methods, then we focus on Monte Carlo dropout for epistemic uncertainty quantification.

\subsection{Uncertainty Evaluation through Bayesian Inference}

To obtain a network capable of providing predictions with an associated level of uncertainty, one possible approach is a Bayesian extension.
In particular, for a regression problem, a traditional neural network implements a predictor $\hat{y} = f(x;w) \in \mathbb{R}$, where $w$ denotes the connection weights.
Bayesian neural networks output a full distribution instead of a point prediction $\hat y$, which is usually modeled with a Gaussian distribution $\mathcal{N}(\hat{\mu}, \hat{\sigma}^2)$, albeit it is not the only possible choice (e.g., Gaussian Mixture distributions or Generalized Gaussian distributions have also been proposed~\cite{Blundell2015,DBLP:conf/eccv/UpadhyayKCMA22}).
In this case, the mean $\hat{\mu}$ is interpreted as the point prediction $\hat{y}$, whereas the variance $\hat{\sigma}^2$ represents the desired uncertainty measure;
ideally, one wants $\hat{\mu}$ and $\hat{\sigma}^2$ to approximate well the \textit{true} target distribution $\mathcal{N}(\mu,\sigma^2)$.

Under a Bayesian framework, the full probability distribution of the target variable $y$ for an input query $x$ is also conditioned on the training set $\mathcal{D}$:
\begin{equation}
    p(y|x,\mathcal{D})=\int_{w} p(y|x,w)p(w|\mathcal{D}) \mathrm{d} w \ .
\end{equation}
This is equivalent to taking the expectation under a posterior on the weights, $p(w|\mathcal{D})$, that provides the desired predictive distribution.
However, since obtaining $p(w|\mathcal{D})$ is intractable for most neural networks, an approximation $q(w)$ is usually considered.
By substituting the true distribution with its approximation, the posterior can be computed as:
\begin{equation}
    p(y|x,\mathcal{D}) \approx \int_{w} p(y|x,w)q(w) \mathrm{d} w \ .
\end{equation}
A suitable approximation $q(w)$ is sought among a set of possible candidates by minimizing its Kullback-Leibler divergence (KL-divergence) $KL[p(w|\mathcal{D})||q(w)]$  with the real distribution.
From variational inference, it is known that the KL-divergence can be minimised by minimizing the negative log-likelihood (NLL), that for regression problems is approximated on a given data set of $N$ samples as: 
\begin{equation}
\label{eqn:neg_log_like}
    \frac{1}{N}\sum_{n=1}^N
    \frac{1}{2}\frac{(y_n-\hat{y}_n)^2}{\hat{\sigma}_n^2}+\frac{1}{2}\log(\hat{\sigma}_n^2) \ .
\end{equation}

\subsection{Monte Carlo Dropout as a Bayesian Approximation}
\label{sec:embedded_dropout}

To model an approximated distribution $q(w)$, one may consider incorporating a stochastic component inside a network. 
One possible way to introduce randomness on the network weights to estimate a full probability distribution over them is to use Stochastic Regularization Techniques~\cite{Gal2016thesis},
originally proposed for network regularisation. 
In particular, it has been shown both theoretically~\cite{GalGahahr16} and practically~\cite{GalGahahr16,Gal2017concretedrop,KendallGal2017} that dropout can be used to this aim.
 
The standard dropout envisages the deactivation of each neuron belonging to a specific network layer with a probability, named \textit{dropout rate}, that follows the Bernoulli distribution $\mathcal{B}(\Phi)$, where the parameter $\Phi$ corresponds to the dropout rate.
The dropout layers are kept active during training and are deactivated at inference time, which produces a regularisation effect.

Unlike standard dropout, Monte Carlo dropout keeps the dropout layers active at inference time; this allows one to derive a full probability distribution $q(w|\Phi)$ approximating $p(w|\mathcal{D})$.
The optimal dropout rate $\Phi$ is the one that minimizes the divergence between the real and the approximated distributions; in practice, a suitable $\Phi$ is found by minimizing an approximation of NLL computed, e.g., on a validation set of $N$ samples:
\begin{equation}
\label{eqn:loqercio_argmin}
    \Phi = \arg \min_{\varphi} \frac{1}{N}\sum_{n=1}^N\frac{1}{2}\frac{(y_n-\hat{y}_n(\varphi))^2}{\hat{\sigma}_n^2(\varphi)}+\frac{1}{2}\log(\hat{\sigma}_n^2(\varphi)) \ .
\end{equation}
Once such a value of $\Phi$ and the trained network are obtained, one can use the approximate distribution $q(w|\Phi)$ for computing a prediction and associated predictive uncertainty, modeled as the mean and the variance of the posterior distribution:
\begin{equation}
\label{eqn:monte_carlo_sampling}
    p(y|x,\mathcal{D}) \approx \frac{1}{T}\sum_{t=1}^{T}p(y|x,w_t) \sim \mathcal{N} (\hat{\mu};\hat{y}, \hat{\sigma}^2) \ ,
\end{equation}
where $T$ denotes the number of Monte Carlo samples obtained by querying the network multiple times with the same sample $x$.

Two widely used techniques for solving the optimization problem~\eqref{eqn:loqercio_argmin} are grid search~\cite{GalGahahr16} and concrete dropout~\cite{Gal2017concretedrop}.
In embedded dropout, after solving~\eqref{eqn:loqercio_argmin} using a training and a validation set, one can use the obtained dropout rate in the inference stage.
To improve performance, a different dropout rate can be used for each network layer~\cite{Gal2017concretedrop}.
On the contrary, in the injected dropout approach~\cite{Loquercio2020} dropout layers are added to an already-trained network, and the dropout rate $\Phi$ is sought by minimizing the NLL~\eqref{eqn:loqercio_argmin} of the trained network, e.g., on a validation set.

\section{The proposed method for Monte Carlo Dropout Injection}
\label{sec:dropout_injection}

A notable difference between embedded and injected dropout is that the former also performs a regularization, whereas the latter does not.
This can lead to differences in the prediction error of the corresponding networks, as well as in their uncertainty measures, which we shall investigate in this section.
In particular, we analyze, theoretically and empirically, how prediction error and quality of the uncertainty measure behave as a function of the dropout rate $\Phi$, for both embedded and injected dropout (Sect.~\ref{sec:effect_dr}).
The results of this analysis will point out an issue of injected dropout, that we propose to mitigate by suitably \emph{scaling} the uncertainty measure; to this aim, we propose an extension of the original formulation in Eq.~\eqref{eqn:loqercio_argmin}, using the $\sigma$-scaling technique~\cite{DBLP:conf/midl/LavesIFKO20} taking into account how the scaling factor affects the calibration of the uncertainty measure(Sect.~\ref{sec:effect_rescaling}).

For simplicity, we shall consider a fixed dropout rate across all network layers, but all our results apply to the most general case when different dropout rates are allowed.

\subsection{Analysis of Monte Carlo Dropout Injection}
\label{sec:effect_dr}

In Sect.~\ref{sec:embedded_dropout} we mentioned that the dropout rate $\Phi$ is usually chosen by minimizing the NLL, both in embedded and injected dropout.
Now, let us consider the weights vector $w$ of a neural network:
one can obtain $w$ by using a standard optimization process that may or may not involve the activation of dropout layers during \emph{training} for regularizing the network weights (i.e., embedded or injected dropout, respectively).
Dropout layers are then activated at \emph{inference} time, both when performing embedded and injected dropout, with dropout rate $\varphi$,
in order to obtain a so-called \textit{Monte-Carlo iteration}, i.e. a prediction taken using the network with stochastic weights deactivation.
This is equivalent to applying a binary mask $m(\varphi) = diag(z_1(\varphi) \dots z_L(\varphi))$ -- where $z_{\ell}(\varphi) \sim \mathcal{B}(\varphi)$ and $L$ denotes the size of $w$ -- to the network weights.
Applying $T$ stochastic iterations is equivalent to applying a series of $T$ masks, each one denoted by $m_t(\varphi)$, to obtain a series of instances of the network weights, $w_1(\varphi) \dots w_T(\varphi)$ with $w_t(\varphi) = w \cdot m_t(\varphi)$, where $\cdot$ denotes the matrix product.
It is easy to see that $w_t(\varphi)$ always depends upon $\varphi$ through the binary mask $m_t(\varphi)$, both for embedded and injected dropout.
We can now rewrite Eq.~(\ref{eqn:monte_carlo_sampling}) for modelling the prediction $\hat{y}$ and the corresponding uncertainty $\hat{\sigma}^2$ of a given instance $x$ when using a neural network with dropout rate $\varphi$:

\begin{equation}
\label{eqn:monte_carlo_sampling_network}
    \frac{1}{T}\sum_{t=1}^{T}p(y|x,w_t(\varphi)) =
    \frac{1}{T}\sum_{t=1}^{T}f(x;w_t(\varphi))
    \sim \mathcal{N} (\hat{\mu};\hat{y}, \hat{\sigma}^2) \ ,
\end{equation}

\noindent where $f(x;w_t(\varphi))$ denotes the prediction of the network parametrized with $w_t(\varphi)$ on the instance $x$.

With abuse of notation, let us denote with $x$, $y$, $\hat{y}$ and $\hat{\sigma}^2$ the \emph{vectors} of the instances, their ground truths, and the corresponding predictions and uncertainty measures, respectively, of a data set $\mathcal{D}$ of size $N$.
It is easy to see that, as the dropout rate changes, the prediction vector $\hat{y} = [\hat{y}_1 \dots \hat{y}_N]$ can change, and as a consequence also the quadratic error vector $\epsilon^2 = (y - \hat{y})^2$ and the uncertainty vector $\hat{\sigma}^2 = [\hat{\sigma}_1^2 \dots \hat{\sigma}_N^2]$ will change.
In particular, choosing the dropout rate $\Phi$ by solving the optimization problem~\eqref{eqn:loqercio_argmin} amounts to pursuing two goals: 
(i) minimizing the quadratic error $\epsilon_n^2 = (y_n - \hat{y}_n)^2$;
(ii) minimizing the absolute difference between the prediction's quadratic error and the variance, $|\epsilon_n^2 - \hat{\sigma}_n^2|$.
It is easy to show that the ``best'' trade-off between them according to Eq.~\eqref{eqn:loqercio_argmin}, i.e., the minimum NLL, is achieved when the following condition holds (a simple proof is reported in~\ref{appendix:A}):
\begin{equation}
    \label{eqn:ideal_uncertainty}
    \hat{\sigma}^2_n = (y_n-\hat{y}_n)^2 \quad \forall n \in \mathcal{D} \ .
\end{equation}

A significant difference can now be noticed in the behavior of $\epsilon^2$ and $\hat{\sigma}^2$, as a function of the dropout rate $\varphi$,
between embedded and injected dropout.
As we already pointed out, when injecting dropout the choice of $\varphi$ does not affect the network weights $w$; in embedded dropout, instead, changing $\varphi$ results in explicitly modifying the vector $w$.
Under the assumption of not having an under-parametrization, the influence of $\varphi$ will lead to a regularization effect on $w$ when using embedded dropout, but not when using injected dropout: since a network with dropout regularization is usually more robust to dropping network weights, it can be expected that a lower quadratic error is attained when using embedded dropout.
Moreover, when using injected dropout we argue that, as $\varphi$ increases, prediction quality quickly deteriorates, i.e., the quadratic error increases.
This behavior penalizes injected dropout, limiting to a tiny range $[0,\varphi_{\rm max}]$, for some $\varphi_{\rm max}$, the values of dropout rate $\varphi$ that lead to acceptably small values of the quadratic error $\epsilon^2$.
However, for very small values of dropout rate, the variability induced on the network predictions, and thus $\hat{\sigma}^2$, are so small that the resulting NLL is very high, which in turn is an indicator of poor quality of the uncertainty measure.
Therefore, the range of ``useful'' values of $\varphi$ further reduces to $[\varphi_{\rm min},\varphi_{\rm max}]$, for some value $\varphi_{\rm min}$. 
We shall provide experimental evidence of this behavior in Sect.~\ref{sec:experiments_crowd}.
In the next section, we propose to address the above issue of injected dropout through a suitable rescaling of its uncertainty measure.

\subsection{The Proposed Injection Method}
\label{sec:effect_rescaling}

The solution we propose to mitigate the above limitation of injected dropout pointed out in Sect.~\ref{sec:effect_dr} is summarized in Fig.~\ref{fig:illustration_injection} and consists in \emph{rescaling} the uncertainty measure $\hat{\sigma}^2$.

\textbf{Rescaling the uncertainty measure --}
Let us consider a rescaled uncertainty measure $\xi^2(\Phi) = C \cdot \hat{\sigma}^2(\Phi)$, where $C \in (0, +\infty)$.
A suitable value of $C$ can be sought by optimizing the NLL as a function of $C$, besides $\Phi$, through the following modification of the optimization problem~\eqref{eqn:loqercio_argmin}:
\begin{equation}
\label{eqn:double_argmin}
    \Phi, C = \arg \min_{\varphi, c} \frac{1}{N}\sum_{n=1}^N\frac{1}{2}\frac{(y_n-\hat{y}_n(\varphi))^2}{c \cdot \hat{\sigma}_n^2(\varphi)}+\frac{1}{2}\log(c \cdot \hat{\sigma}_n^2(\varphi))\ .
\end{equation}

\begin{figure}[tb]
    \centering
    \includegraphics[width=\textwidth]{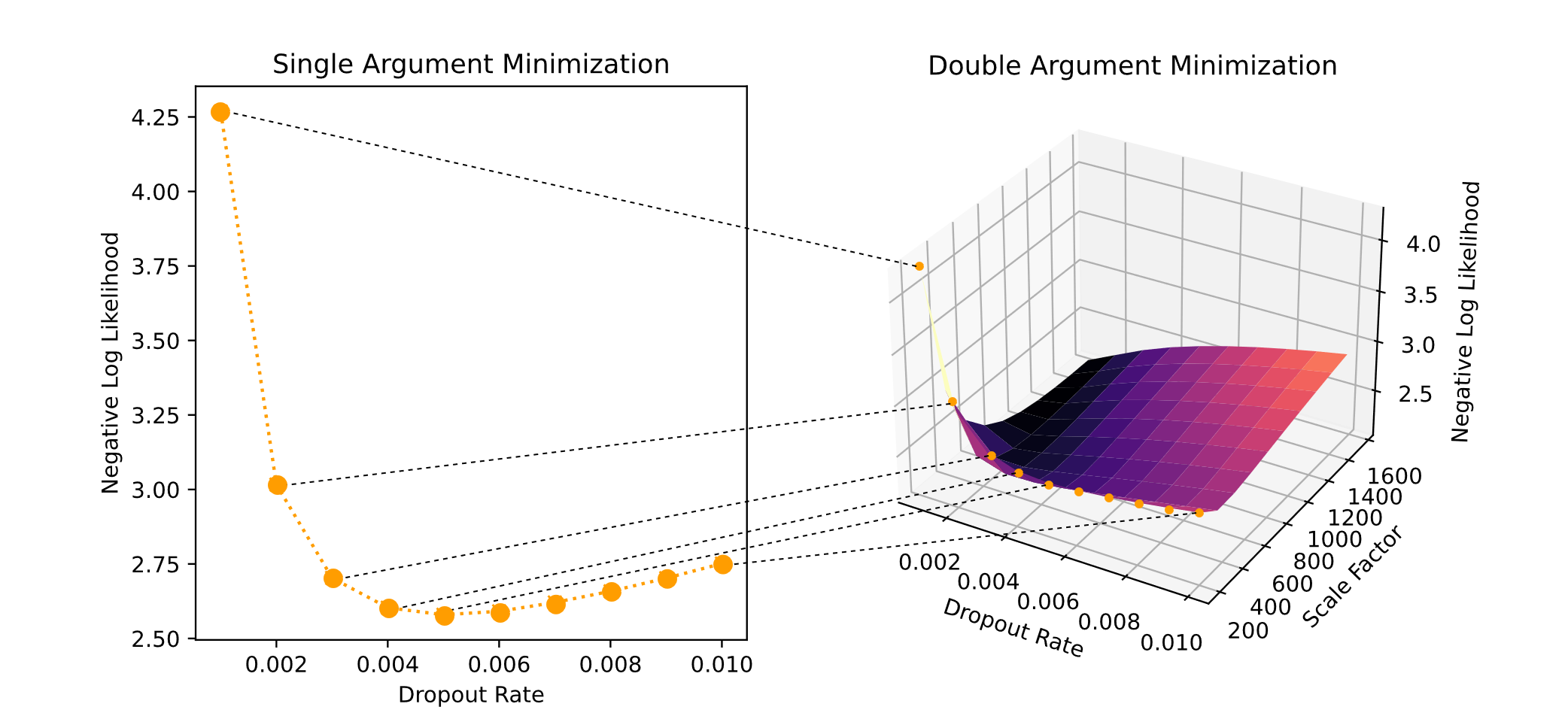}
    \caption{These plots (taken from experiments in Sect.~\ref{sec:experiments_crowd}) point out the difference between the scale-agnostic optimization process of Eq.~\eqref{eqn:loqercio_argmin} (left) and the scale-aware one of Eq.~\eqref{eqn:double_argmin} (right). The latter can potentially lead to different dropout rates with a lower NLL than the former, thanks to a rescaling of the uncertainty measure.}
    \label{fig:optimization_surface}
\end{figure}

We argue that such a scaled measure can lead to a much better NLL value than the original formulation~\eqref{eqn:loqercio_argmin}.
To solve the optimization problem~\eqref{eqn:double_argmin} it is possible to exploit the results by Laves et al.~\cite{DBLP:conf/midl/LavesIFKO20}, which show that, given a trained regressor providing a prediction $\hat{y}$ and a corresponding uncertainty measure $\hat{\sigma}^2$, one can further reduce the NLL with respect to the solution of~\eqref{eqn:neg_log_like} by rescaling $\hat{\sigma}^2$ as $C \cdot \hat{\sigma}^2$, and that the optimal value of the scale factor $C$, for any \emph{given} quadratic error and uncertainty measure vectors $\epsilon_n^2$ and $\hat{\sigma}^2_n$, is analytically given by: 
\begin{equation}
    \label{eqn:laves}
    C = \frac{1}{N}  \sum_{n=1}^N\frac{\epsilon_n^2}{\hat{\sigma}^2_n} .
\end{equation}
Accordingly, the optimal scale factor can be interpreted as an intrinsic indicator of how much the magnitude of the uncertainty vector is close to the magnitude of the quadratic error: if $C>1$, the uncertainty measure $\hat{\sigma}_n$ is smaller, on average (over the $N$ samples), than the quadratic error $\epsilon_n$; on the contrary, if $C<1$, then $\hat{\sigma}_n$ is on average larger than $\epsilon_n$.
Finally, since the scale factor is analytically known for a given uncertainty measure, Eq.~\eqref{eqn:laves} can be exploited to rewrite $C$ as a function of $\Phi$ in the optimization problem~\eqref{eqn:double_argmin}, which can therefore be simplified as:
\begin{equation}
\label{eqn:single_argmin}
    \Phi = \arg \min_{\varphi} \frac{1}{N}\sum_{n=1}^N\frac{1}{2}\frac{(y_n-\hat{y}_n(\varphi))^2}{C(\varphi) \cdot \hat{\sigma}_n^2(\varphi)}+\frac{1}{2}\log(C(\varphi) \cdot \hat{\sigma}_n^2(\varphi)) \ .
\end{equation}
Accordingly, both the dropout rate and the scale factor can be found by optimizing the NLL only with respect to the dropout rate, similarly to the original formulation~\eqref{eqn:loqercio_argmin}.

When introducing a scale factor, the values of the unscaled uncertainty measure $\hat{\sigma}_n^2$ do not need anymore to be close to the corresponding quadratic error $\epsilon^2_n$ to keep the NLL small, and are therefore not penalized as in the original formulation~\eqref{eqn:loqercio_argmin}. 
We point out that, although rescaling can also be applied to embedded dropout, it is much more advantageous for injected dropout to mitigate the issue described in Sect.~\ref{sec:effect_dr}, which instead does not affect embedded dropout to a significant extent.
In particular, such mitigation is expected to be more beneficial for large data sets, complex tasks, and networks with a large number of weights: we shall provide experimental evidence of this fact in Sect.~\ref{sec:experiments_crowd}.
As shown in Fig.~\ref{fig:optimization_surface}, the proposed formulation~\eqref{eqn:single_argmin} potentially allows one to explore regions of the NLL surface (considered as a function of both the dropout rate and the scale factor), that would otherwise be inaccessible by the original formulation~\eqref{eqn:loqercio_argmin}, leading to a better NLL.

\textbf{Calibration of the uncertainty measure as a function of the scale factor --}
So far, we have focused on NLL minimization, since this allows one to minimize the distance between the actual prior distribution of the weights, $p(w)$, and the approximated one, $q(w)$.
However, the quality of an uncertainty measure also depends on its \emph{calibration}, and minimizing NLL does not guarantee to obtain a well-calibrated measure.
Given the relevance of the scaling factor to injected dropout, it is interesting to investigate how it affects the trade-off between minimizing NLL and obtaining a well-calibrated measure.
To this aim, we first summarise the notion of calibration of an uncertainty measure for regression problems.
As mentioned in Sect.~\ref{sec:statement}, the distribution
$p(y|x,\mathcal{D})$ is assumed to be Normal, $\mathcal{N}(\mu, \sigma^2)$, as well as its approximation in Eq.~\eqref{eqn:monte_carlo_sampling},
$\mathcal{N}(\hat{\mu}, \hat{\sigma}^2)$; to make the latter as close as possible to the former, one minimizes the NLL on a validation set.
Considering $\mu$ and $\sigma^2$ as the point prediction and the corresponding uncertainty for a given instance $(x,y)$, one can construct a confidence interval with probability $\alpha$ for a generic sample $(x,y)$:
\begin{equation}
CI_{\alpha} = [\mu - z_{\alpha} \cdot \sigma^2, \mu + z_{\alpha} \cdot \sigma^2] \ ,
\end{equation}
which is characterized by the following:
\begin{equation}
\label{eqn:conf_interval}
    p(y \in CI_{\alpha})=\alpha, \quad \forall \alpha \in [0,1] \ .
\end{equation}
One would like that Eq.~\eqref{eqn:conf_interval} holds for any instance $(x,y)$ also with respect to a confidence interval $\hat{CI}_{\alpha}$ computed from the estimated distribution:
\begin{equation}
\hat{CI}_{\alpha} = [\hat{\mu} - z_{\alpha} \cdot \hat{\sigma}^2, \hat{\mu} + z_{\alpha} \cdot \hat{\sigma}^2] \ .
\end{equation}
If so, the uncertainty measure $\hat{\sigma}^2$ is said to be perfectly calibrated.
This is, however, not guaranteed in practice.

\begin{figure}
    \centering
    \includegraphics[width=0.5\textwidth]{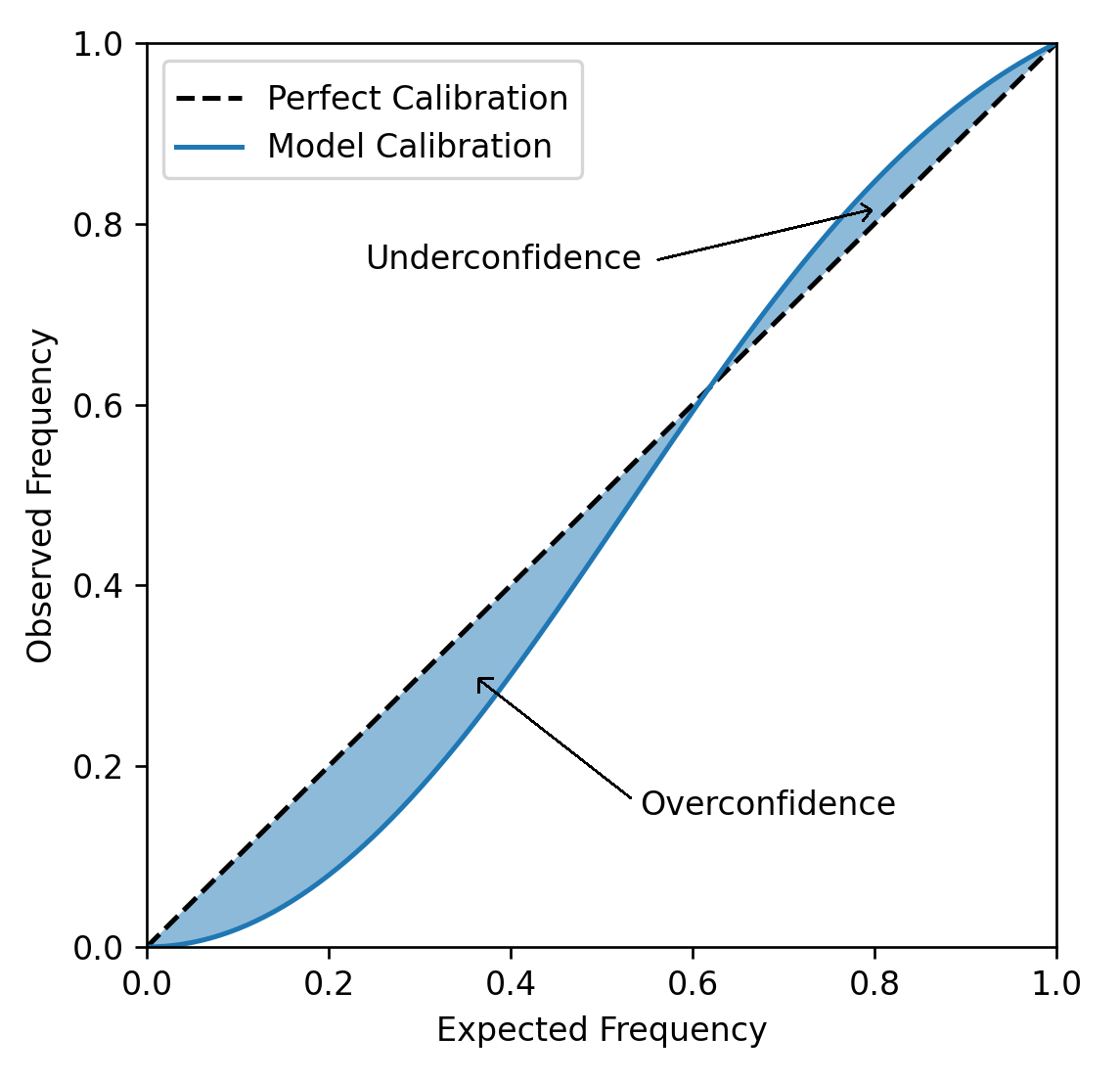}
    \caption{Example of a calibration curve, which plots the observed frequency of instances belonging to the confidence interval $CI_\alpha$ with probability $\alpha$ as a function of the respective expected frequency. In this example, an under-confidence and lower-confidence regions are present (see the text for their meaning).}
    \label{fig:reliability_diagram_explained}
\end{figure}

To empirically evaluate the calibration of an uncertainty measure, the probability $p(y \in CI_{\alpha})$ can be estimated on a data set (referred to as validation~\cite{DBLP:conf/icml/GuoPSW17} or calibration set~\cite{DBLP:conf/icml/KuleshovFE18}), as the fraction of instances belonging to the confidence interval $\hat{CI}_{\alpha}$, for several values of $\alpha$.
We call this fraction the \emph{observed frequency}, and denote it by $\pi(\alpha)$; ideally, $\pi(\alpha)$ should be equal to $\alpha$, which we call the \emph{expected frequency}.
Calibration quality can therefore be evaluated through a \textit{calibration curve}, i.e., a plot of the observed frequency as a function of the expected one, as in the example shown in Fig.~\ref{fig:reliability_diagram_explained}. 
Note that a perfect calibration corresponds to a calibration curve equal to the bisector of the X-Y axes.
When the observed frequency $\pi(\alpha)$ is below the bisector, the uncertainty measure is \emph{overconfident} for the corresponding values of $\alpha$ (because the expected frequency is greater than the observed one); on the contrary, where $\pi(\alpha)$ is above the bisector, the uncertainty measure is \emph{underconfident}.\footnote{Note that the calibration curve can cross the bisector for zero, one or more points, besides the ones corresponding to $\alpha=0$ and $\alpha=1$.}

Calibration quality can now be quantitatively evaluated as a scalar measure, in terms of ``how far'' the calibration curve is from the ideal one.
In this work, we employ the area subtended by the calibration curve with respect to the bisector, known in the literature by the name of Miscalibration Area (MA)~\cite{DBLP:journals/mlst/TranNYZXU20}, which corresponds to the shaded regions in Fig.~\ref{fig:reliability_diagram_explained}:
\begin{equation}
\label{eqn:analitic_miscalibration_area}
{\rm MA} = \int_0^1 |\pi(\alpha) - \alpha| {\rm d}\alpha \ . 
\end{equation}
In practice, the MA can be empirically estimated from a validation set as the average difference in absolute value between the expected and the observed frequency, over a set of $M$ values of $\alpha$:
\begin{equation}
    \label{eqn:area_under_calib_curve}
    {\rm MA} = \frac{1}{M} \sum_{m=1}^M |\pi(\alpha_m) - \alpha_m| \ .
\end{equation}

As mentioned above, the scale factor $C$ obtained by solving problem~\eqref{eqn:single_argmin} does not guarantee that the corresponding MA is null, i.e., perfect calibration. 
To analyze the trade-off between MA and NLL that can be obtained as a function of the scale factor, we devised an iterative procedure that, starting from the obtained scale factor $C$, modifies it to reduce MA as much as possible, by increasing or reducing the confidence of each prediction of the calibration set, depending on whether the uncertainty measure is over- or under-confident on that prediction, respectively.

Our procedure (reported as Algorithm~\ref{algo:relaxed_scale_factor}) aims at reducing to zero the average difference between the expected and the observed frequency (which we refer to as $Balance$) as a proxy for minimizing the MA.
In such a setting, the value of $Balance$ depends upon the chosen scale factor $c$; we denote as $Balance_c$ such dependence, which is also reflected on the observed frequency $\pi_c(\alpha)$.
Since $Balance_c$ is a monotonically non-decreasing function of $c$, one can empirically construct a suitable interval $[C_{\rm low}, C_{\rm high}]$ around the scaling factor $C$ obtained by solving problem~\eqref{eqn:single_argmin}, such that  $Balance_{C_{\rm low}}$ is negative, and $Balance_{C_{\rm high}}$ is positive;
then, our algorithm uses the bisection method to find a ``relaxed'' scaling factor $C_{\rm r}$ such that $Balance_{C_{\rm r}}=0$ (in practice, the objective is defined as $Balance_{C_{\rm r}} < \tau$, where $\tau$ is a given tolerance threshold).

By analyzing the ``path'' of the scale factor between $C$ and $C_{\rm r}$, i.e., the corresponding values of NLL and ME, it is possible to explore the trade-off between these two measures.
Although acting on the scale factor does not guarantee to reduce MA to arbitrarily small values (i.e., to approach perfect calibration), empirical results reported in Sect.~\ref{sec:experiments} show that small enough MA values can nevertheless be obtained.
Interestingly, our results also show that the corresponding increase of NLL can be very limited:
therefore, although standard post-hoc techniques can be applied (e.g., Kuleshov calibration~\cite{DBLP:conf/icml/KuleshovFE18}) to enhance the calibration quality,
the above result suggests that, when using injected dropout, a good trade-off between minimizing NLL and obtaining a well-calibrated uncertainty measure can also be achieved by first computing the dropout rate and the corresponding scale factor through problem~\eqref{eqn:single_argmin}, and then suitably adjusting the latter.

\algdef{SE}[DOWHILE]{Do}{doWhile}{\algorithmicdo}[1]{\algorithmicwhile\ #1}%

\begin{algorithm}[tb]
\caption{Modifying the scale factor to improve calibration}
\label{algo:relaxed_scale_factor}
\begin{algorithmic}
\Require vectors of uncertainty measures $\hat{\sigma}^2$, network predictions $\hat{y}$ and ground truth $y$ for a data set of $N$ instances; a suitable interval $[C_{\rm low}, C_{\rm high}]$ around the initial scaling factor $C$; $M$ values $\alpha_1, \ldots, \alpha_M$; tolerance threshold $\tau > 0$ 
\Ensure the relaxed scaling factor $C_{\rm r}$
\Do
    \State $C_{\rm r} \gets (C_{\rm low}+C_{\rm high} )/ 2$
    \State $Balance_{C_{\rm r}} \gets \frac{1}{M} \sum_{m=1}^M (\pi_{C_{\rm r}}(\alpha_m) - \alpha_m)$ 
    \If{$Balance_{C_{\rm r}} < 0$}
        \State $C_{\rm low} \gets C_{\rm r}$
    \ElsIf{$Balance_{C_{\rm r}} > 0$}
        \State $C_{\rm high} \gets C_{\rm r}$
    \EndIf
\doWhile{$Balance_{C_{\rm r}} \ge \tau$}

\end{algorithmic}
\end{algorithm}

\section{Experiments}
\label{sec:experiments}

We carried out two groups of experiments.
We first compared the behavior of embedded and injected dropout on eight data sets related to regression problems taken from the UCI repository\footnote{\url{https://archive.ics.uci.edu/ml/datasets.php}}, 
using a multi-layer feed-forward network as the regression model.
In the second group of experiments, we evaluated injected dropout on five benchmark data sets for crowd counting~\cite{Oh2020}, which is a challenging computer vision application using a state-of-the-art CNN architecture for this task.
In both experiments, we implemented injected dropout by setting the dropout rate and scaling factor according to Eq.~\eqref{eqn:single_argmin}, using a validation set; in the experiments on crowd counting we also considered the relaxed scaling factor obtained as described in Sect.~\ref{sec:effect_rescaling} on the same validation set.

We evaluated both embedded and injected dropout, as a function of the dropout rate, in terms of the three metrics considered in Sect.~\ref{sec:dropout_injection}: root mean squared error (RMSE), NLL, and MA.
In particular, we evaluated NLL and MA using both the unscaled values $\hat{\sigma}^2(\varphi)$ and the scaled ones $\xi^{2}(\varphi)$; for reference, we also computed the ``ideal'' uncertainty measure of Eq.~\eqref{eqn:ideal_uncertainty}, defined as the error vector $\epsilon^2$ computed on the \emph{validation set}

\subsection{Experiments on UCI Data Sets}

We selected eight data sets related to several multivariate regression problems that have been used in the literature to evaluate uncertainty measures:
Protein-Tertiary-Structure, Boston-Housing, Concrete, Energy, Kin8nm, Power-Plant, Wine-Quality-Red, and Yacht.
We adopted the same set-up proposed by Hernandez-Lobato et al.~\cite{HernandezLobato}, described in the following.
We also open-sourced the code for the UCI experiment reproducibility at the following link \url{https://github.com/EmanueleLedda97/Dropout_Injection}.

\textbf{Experimental set-up --}
We first distinguished between the challenging Protein-Tertiary-Structure data set and the remaining simpler ones.
We used a one-hidden-layer network with 50 hidden units for the latter and 100 for the former.
We randomly split each data set into a training set made up of 90\% of the original instances and a testing set containing the remaining 10\%.
This procedure is repeated five times for Protein-Tertiary-Structure and 20 times for the other data sets.
In each split, we used 20\% of the training partition as a validation set, which is crucial not only for proper hyper-parameter tuning but also -- and primarily -- for scaling the uncertainty measure.

Differently from the original set-up~\cite{HernandezLobato}, we used a mini-batch stochastic gradient descent; we empirically tuned the batch size, the learning rate, and the number of epochs on the validation set.
We considered $15$ possible dropout rates within the interval $[0.001, 0.5]$. We chose these rather extreme values to properly investigate the issue mentioned in Sect.~\ref{sec:effect_rescaling}, where we hypothesized a significant deterioration of the NLL of the unscaled measure for such values of dropout rate.

\begin{figure*}[tb]
    \centering
    \includegraphics[width=\textwidth]{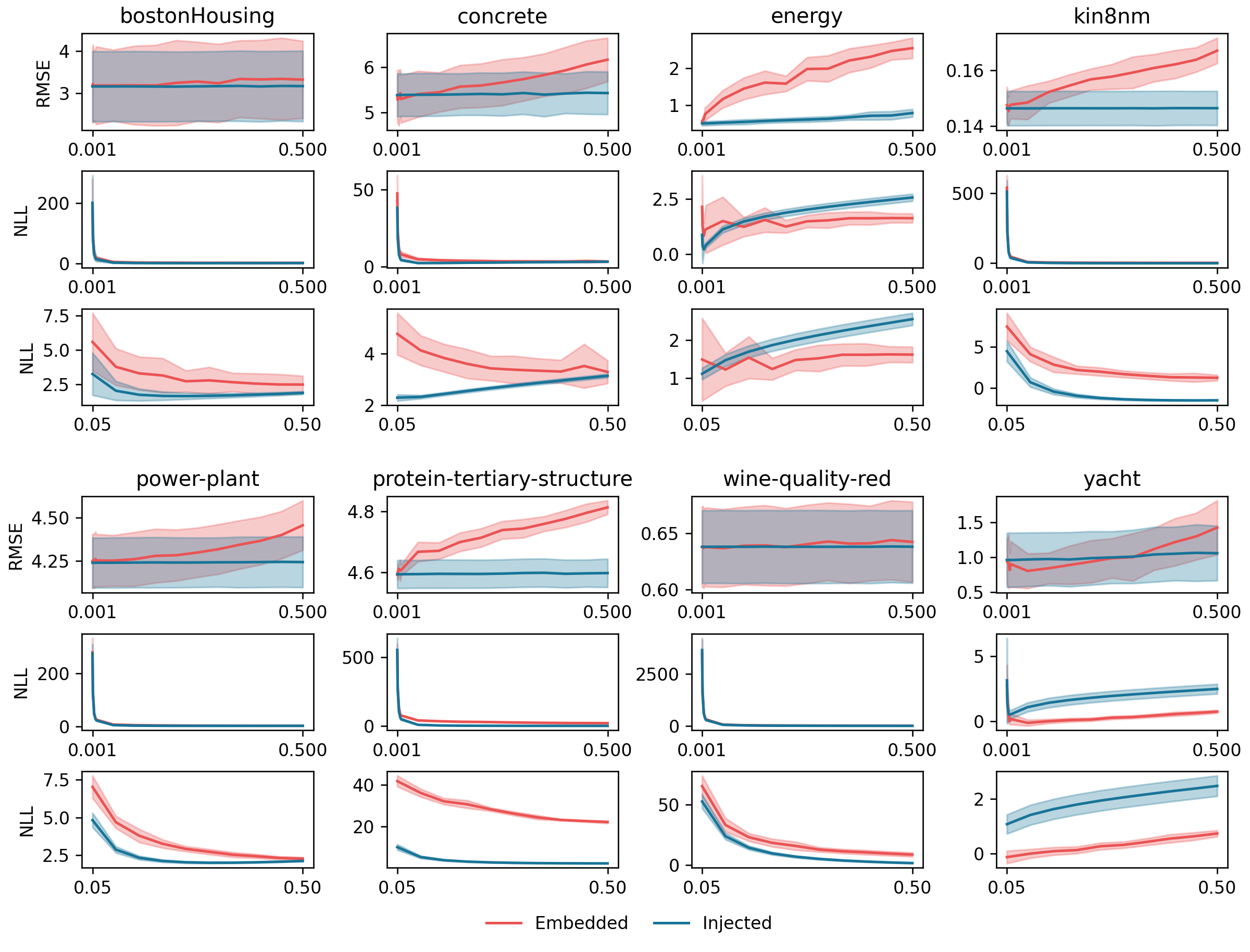}
    \caption{RMSE and NLL as a function of dropout rate on UCI data sets, for injected (blue) and embedded (red) dropout. Two NLL plots are reported for each data set, related to different ranges of dropout rate: $[0.001, 0.5]$ (top plot, same range as the RMSE plot) and $[0.05, 0.5]$ (bottom plot), to better visualize the Y-axis scale.} 
    \label{fig:uci_plot1}
\end{figure*}

\textbf{Experimental results --}
Results are reported in Fig.~\ref{fig:uci_plot1}.

\noindent We can first notice that the RMSE of embedded dropout tends to deteriorate as the dropout rate increases.
This behavior suggests that the network's capacity is very low and that the regularization effect of dropout leads to under-fitting.
This over-penalization makes injected dropout more accurate on average, in terms of average RMSE and standard deviation, although a different behavior can be observed on different data sets.
Another phenomenon we can observe is that, for the smallest dropout rates, the RMSE is low, but the unscaled NLL is almost always severely affected.
This trend reveals that -- as we hypothesized in Sect.~\ref{sec:effect_rescaling} -- unscaled uncertainty measures suffer from the correlation between the dropout rate and the corresponding network perturbation's magnitude.
When the dropout rate is extremely low, the final perturbation (and therefore the uncertainty measure) is so small that it completely disagrees with the quadratic prediction error, worsening the NLL.

\begin{figure*}[tb]
    \centering
\includegraphics[width=\textwidth]{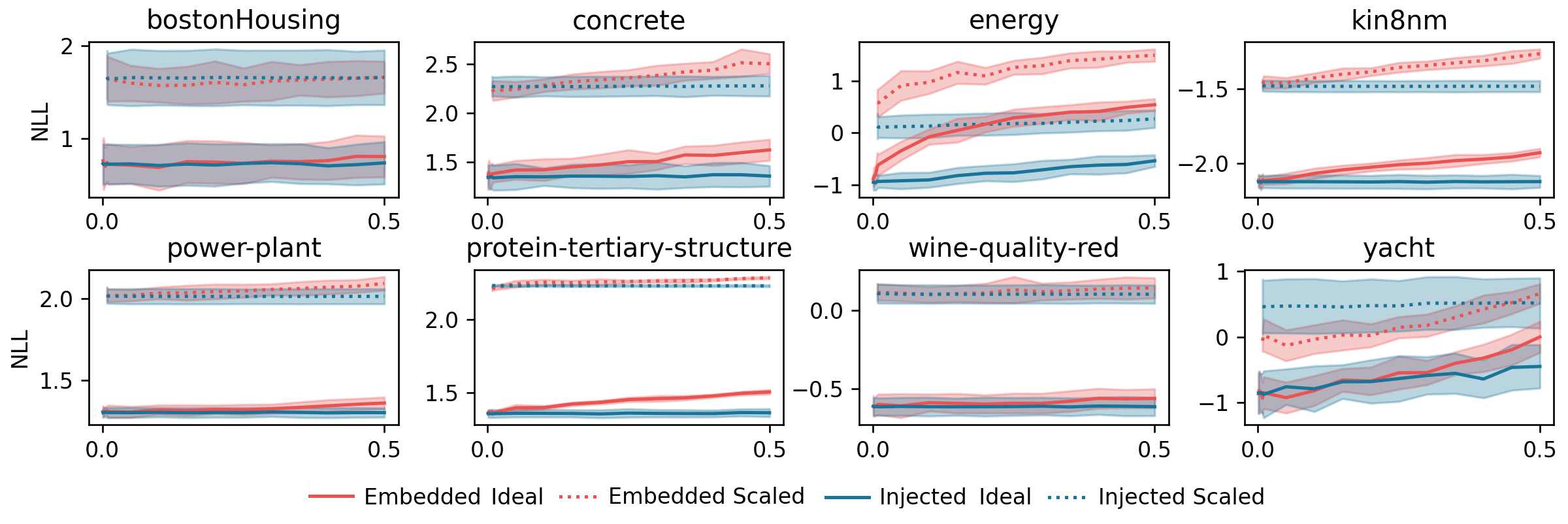}
    \caption{NLL as a function of dropout rate on UCI data sets, for the scaled (dotted line) and ideal (solid line) uncertainty measures, both for injected (blue) and embedded (red) dropout.}
    \label{fig:uci_plot2}
\end{figure*}

Fig.~\ref{fig:uci_plot2} shows that the scaled measures exhibit significantly different behavior instead.
In this case, it is evident that even for extremely low dropout rates, the NLL does not deteriorate as much as the unscaled one, in line with our hypothesis in Sect.~\ref{sec:effect_rescaling}. 
Although the observed trends are always in favor of the injected dropout, it may be possible that this behavior was due to the use of a multi-layer network with a hidden layer of 50 or 100 units, and -- as we said before -- there is clear evidence that using the dropout at training time in this set-up leads to under-fitting the considered data sets. 
This deterioration is also visible by looking at the ideal NLL: the solid red line in Fig.~\ref{fig:uci_plot2} tends to be higher than the solid blue line, which means that the ideal uncertainty measure (which is equal to the prediction error, for each sample) is already penalized.
Therefore, the observed phenomena are entirely in line with our theoretical considerations in Sect.~\ref{sec:effect_rescaling}.

\begin{figure*}[tb]
    \centering
    \includegraphics[width=\textwidth]{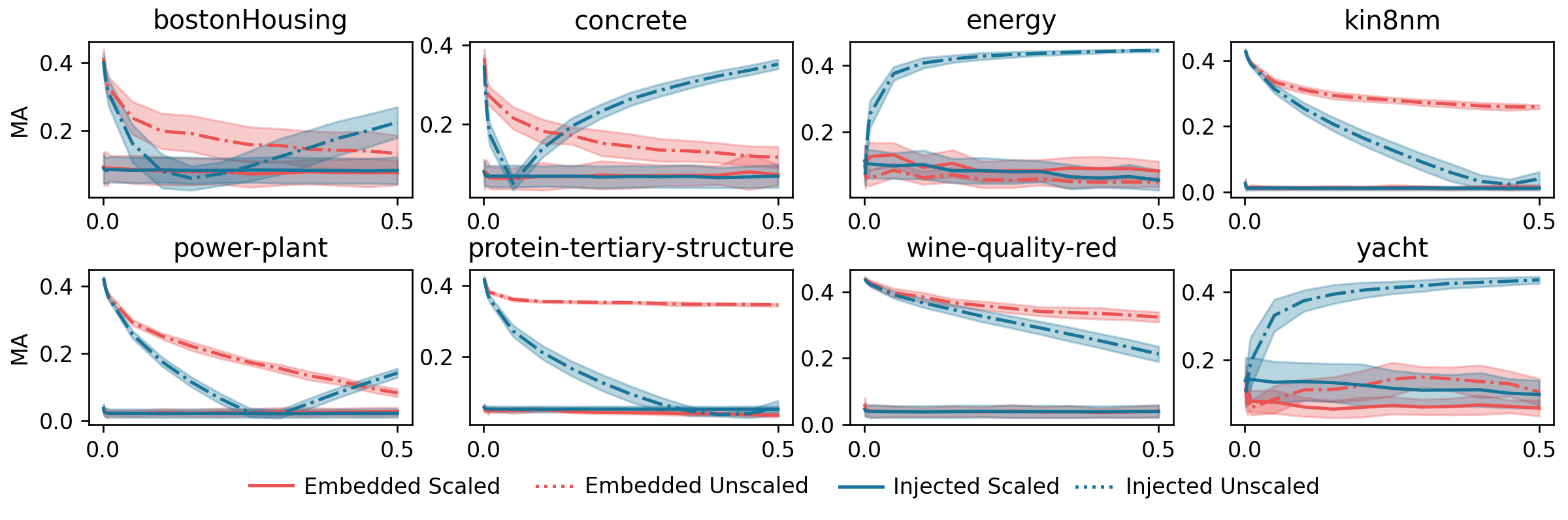}
    \caption{Calibration error as a function of dropout rate on UCI data sets, using the unscaled (dotted line) and scaled (solid line) uncertainty measures, both for injected (blue) and embedded (red) dropout.}
    \label{fig:uci_plot3}
\end{figure*}

Looking at the calibration error in Fig.~\ref{fig:uci_plot3}, we can see that dropout injection can provide well-calibrated uncertainty measurements.
Analyzing the trends of the calibration error in detail, we can notice the following behaviors: first of all, the unscaled measures are clearly more sensitive to changes in dropout rate compared to their scaled counterparts.
However, although the scaled measures seem much more stable, there are cases where the unscaled measure of embedded dropout performs comparably to or even outperforms the scaled one; this phenomenon occurs more sporadically -- and much less clearly -- for injected dropout.
This is a crucial issue if we consider that, for the corresponding dropout rates, a trade-off can be observed between RMSE, NLL, and MA; this suggests that the embedded dropout inherently achieves a trade-off between these metrics.
A further investigation of this issue is an interesting direction for future work.

\subsection{Experiments on Crowd Counting}
\label{sec:experiments_crowd}

The experiments done on UCI data sets provide useful insights about the behavior of injected vs. embedded dropout and evidence that also the former can be an effective approach to epistemic uncertainty quantification.
In the next experiments, we focused on a much more challenging computer vision problem, i.e., crowd counting.
State-of-the-art CNN models solve it by first estimating the crowd density map, which is an image-to-image regression problem.

\textbf{Experimental set-up --}
We used three benchmark crowd counting data sets, UCSD~\cite{UCSDDS}, Mall~\cite{MallDS}, and PETS~\cite{Ferryman2009PETS2009DA}. 
For the latter, we used three different partitions corresponding to different crowd scenes, named PETSview1, PETSview2, and PETSview3~\cite{Delussu2020}.
We used the Multi-Column Neural Network architecture~\cite{DBLP:conf/cvpr/ZhangZCGM16}, a well-known state-of-the-art CNN for crowd counting. We implemented dropout injection right before each convolutional layer, except for the first one, following the suggestion of previous works on Monte Carlo dropout~\cite{Gal2017concretedrop}.
Unlike the previous experiments, we found an exponential increase in the RMSE using the same range of dropout rates.
Therefore, we followed the suggestion by Loquercio et al.~\cite{Loquercio2020} to use very small dropout rates, which leads to a quasi-deterministic network. In particular, we considered ten dropout rates in the interval $[0.001, 0.01]$.

\begin{figure*}[tb]
    \centering
    \includegraphics[width=\textwidth]{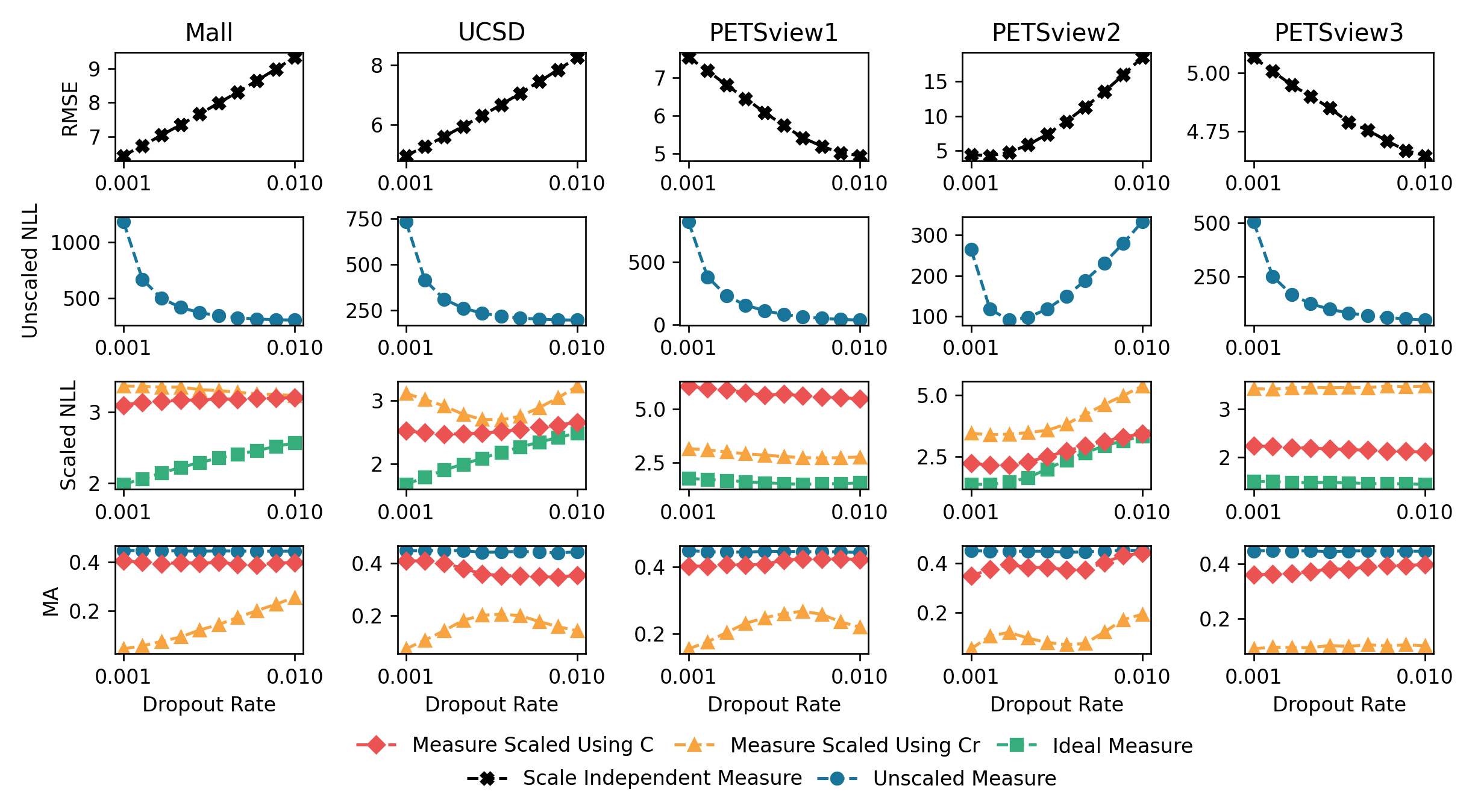}
    \caption{RMSE, NLL, and calibration error of the injected dropout as a function of dropout rate for the crowd counting data sets. Two plots are reported for NLL, corresponding to the unscaled (top plot) and scaled (bottom plot) uncertainty measure. The plots of scaled NLL and calibration error report also a comparison with the ideal scale (the former) and with the unscaled measure (the latter)} 
    \label{fig:crowd_plots}
\end{figure*}

\textbf{Experimental results --}
\noindent Fig.~\ref{fig:crowd_plots} reports the RMSE, NLL (scaled and unscaled), and MA as a function of dropout rate.
Since there is a significant difference in the magnitude of the NLL for the unscaled measures and the scaled ones, we reported them into two different plots.

We can observe an increase in RMSE in three of five data sets, while for two of them (PETSview1 and PETSview3), the RMSE decreased.
This apparently different behavior is due to the very small dropout rates used in these experiments; we observed indeed that for higher dropout rates, the RMSE started to increase also for PETSview1 and PETSview3.

Regarding the NLL, a large difference between the scaled and the unscaled measures can be observed, even in this experiment: similarly to the UCI data sets, the smallest dropout rates lead to a poor unscaled measure, whereas the scaled measure attained considerably better results.
It is also possible to see how the best dropout rate values differ between scaled and unscaled measures: for example, in UCSD, the minimum NLL is around $\Phi=0.04$ for the scaled measure whereas, for the unscaled measure, it is around $\Phi=0.01$.
Similar considerations can be made for Mall.
We can also notice that the scaled measure is much more stable than the unscaled one, even with the smallest dropout rates: this can be advantageous when using injected dropout, as it allows one to make minimal changes to the network to obtain suitable measures of uncertainty.
Although re-scaling performs well in terms of the NLL, the same does not hold for the calibration error: in this case, sub-optimal performances can be observed, with -- in most cases -- a miscalibration area greater than $0.3$, which is an indicator of poor calibration.
Therefore, in this case, a suitable trade-off between NLL and calibration error should be sought, e.g., by tuning the scaling factor obtained from Eq.~\eqref{eqn:single_argmin} through the relaxation procedure presented in Sect.~\ref{sec:effect_rescaling}.

\begin{figure*}[tb]
    \centering
    \includegraphics[width=\textwidth]{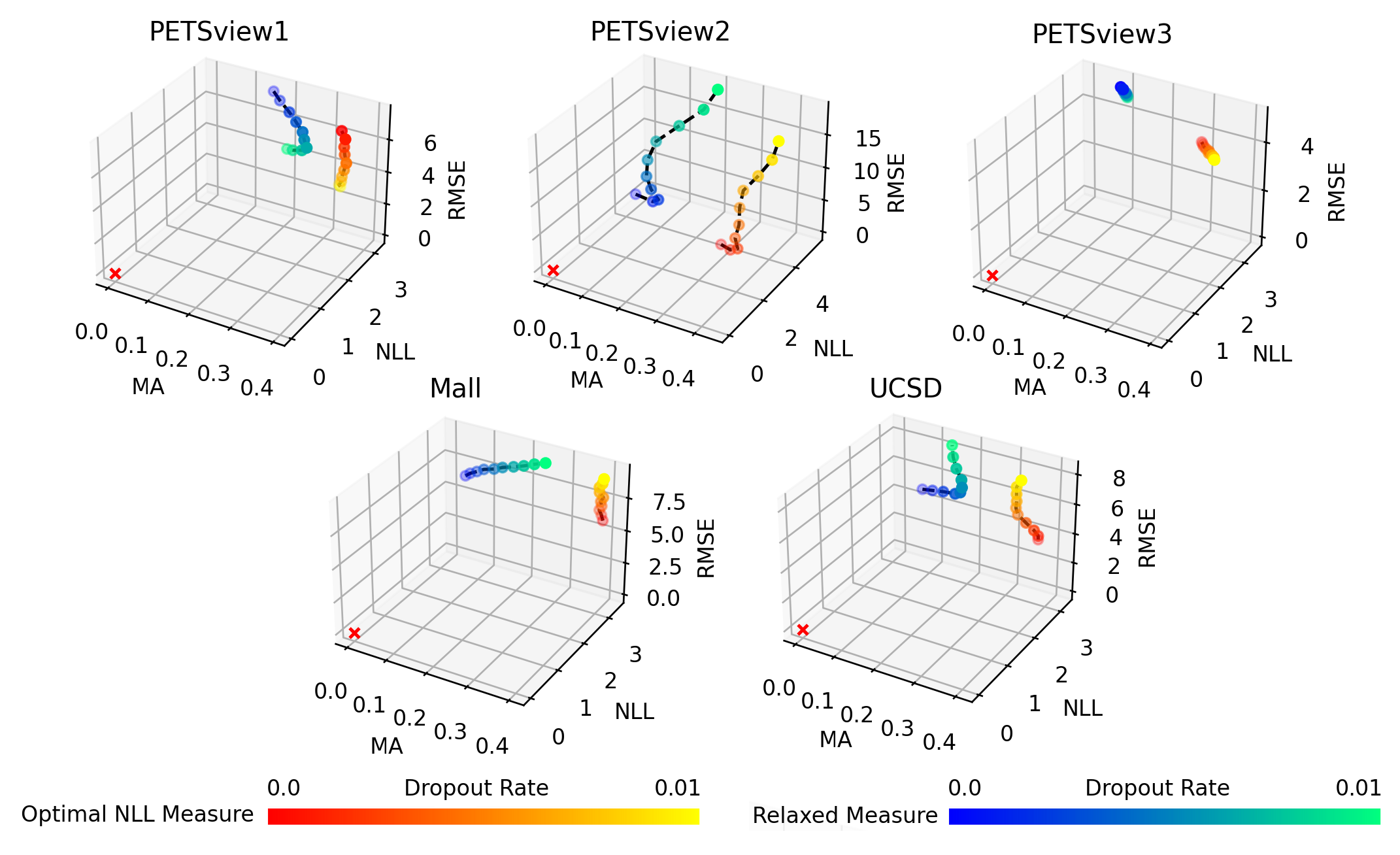}
    \caption{Trade-off between RMSE, NLL, and MA of the injected dropout, as a function of dropout rate, on the crowd counting data sets, for the optimal scaling (warm color gradient) and the relaxed one (cold color gradient).}
    \label{fig:3dplot_tradeoff}
\end{figure*}

Using a relaxed version of the scaling factor allows one to attain a trade-off between RMSE, NLL, and MA, instead of minimizing only the NLL, by manually tuning the scaling factor;
Fig.~\ref{fig:3dplot_tradeoff} shows the value of the three analyzed metrics as a function of dropout when using both $C$ (cold-colored gradient) and $C_{\rm r}$ (warm-colored gradient).
It is easy to notice that when using $C_r$, the NLL is not far from the values obtained using $C$: this suggests that the relaxed measure behaves as a reasonable compromise between NLL and MA.

\begin{figure*}[tb]
    \centering
    \includegraphics[width=\textwidth]{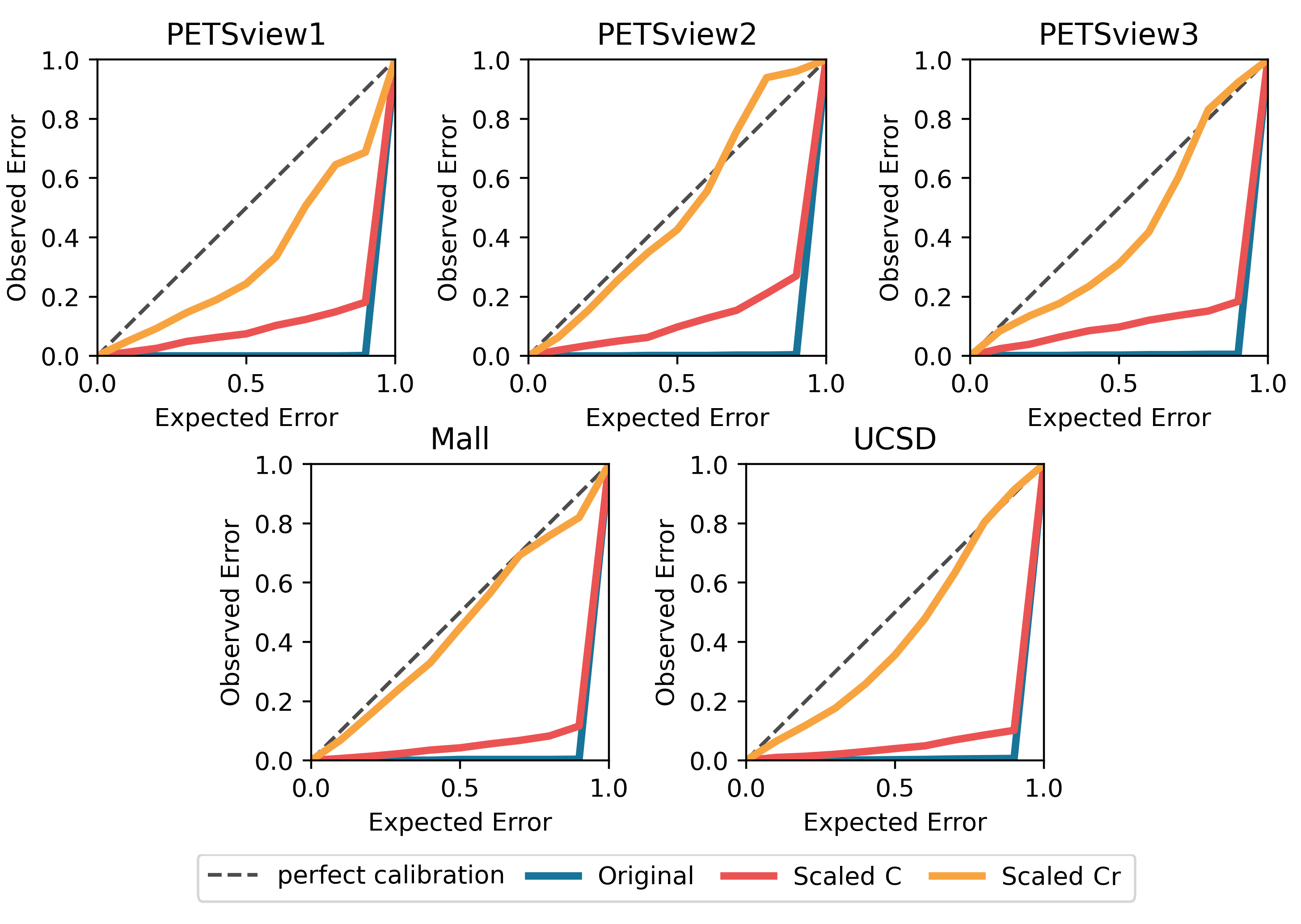}
    \caption{Calibration curves of injected dropout on the crowd counting data sets, attained using the unscaled measure (blue), the optimal scaling factor $C$ (red), the relaxation $C_r$ (orange). For reference, the perfect calibration line is also shown (dotted line).}
    \label{fig:calibration_curves}
\end{figure*}

Finally, the calibration curves in Fig.~\ref{fig:calibration_curves} clearly show that the proposed relaxation procedure effectively reduces the calibration error. In all cases, the relaxed curve is much closer to the axes' bisector, supporting our hypothesis in Sect.~\ref{sec:effect_rescaling}. 

\textbf{Practical utility of uncertainty quantification by dropout injection in video surveillance applications --}
We first show how the operator of a video surveillance system can benefit from updating its crowd density estimator software by integrating an uncertainty estimate through dropout injection (i.e., without requiring further training), and then how uncertainty evaluation could be automatically exploited to improve the accuracy of the estimated density maps and of the corresponding people count.

\begin{figure}
    \centering
    \includegraphics[width=\textwidth]{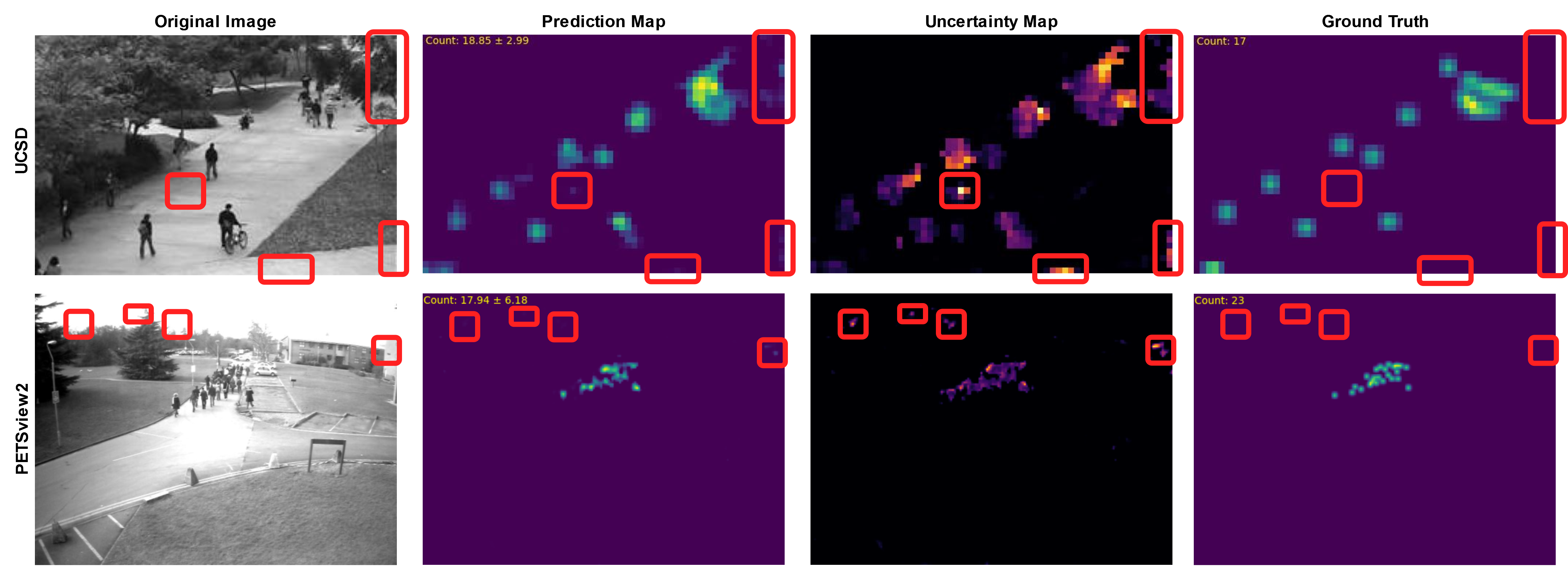}
    \caption{Two examples of ``false positive'' detections enabled by uncertainty estimation on crowd-counting images (top: UCSD, bottom: PETSview2). From left to right: the original frame, ground truth, predicted density map, and uncertainty map. Each density map is normalized: darker pixels on the gradient palette (\img{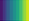} for ground truth and predicted maps, \img{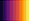} for the uncertainty maps) correspond to lower values. ``False positive'' regions, corresponding to low predicted density and high uncertainty are highlighted with red boxes.}
    \label{fig:highlighted_uncertainty_maps}
\end{figure}

Crowd counting by density estimation uses density maps as ground truth: the standard method for obtaining such maps starts by annotating each image location where one person's head is present,
then the density map is obtained by adding a kernel (e.g., Gaussian) with a unit area centered on each head location. Therefore, the people count corresponds to the sum of each pixel of the density map.
In this context, a crowd counting predictor will output an estimated density map reconstructing the original one.
The image-to-image nature of this task, similar to what happens in Monocular Depth Estimation and Semantic Segmentation, implies that also the uncertainty estimate consists of a 2D array, associating a level of uncertainty to each pixel-wise predicted density (i.e., its variance).
In turn, the pixel-wise uncertainty on the predicted density allows the construction of a confidence interval on the people count, as seen in the examples in Fig.~\ref{fig:highlighted_uncertainty_maps}, similarly to \cite{Oh2020}, but in our case, in a post hoc fashion.
Having a notion of the confidence level associated with system predictions makes video surveillance operators aware of their reliability, allowing them to make more informed decisions and increasing their trust in the prediction system compared to a point estimator.

An in-depth analysis of our experimental results on crowd counting benchmarks reveals that uncertainty evaluation could also be helpful to detect some inaccuracies in the estimated density maps that correspond to ``false positive'' pedestrian localization, i.e., image regions where the predicted density is not zero (thus contributing to the pedestrian count) despite the absence of pedestrians.
As seen from the examples in Fig.~\ref{fig:highlighted_uncertainty_maps}, we observed that false positive localizations are often characterized by isolated regions with relatively low estimated density and high uncertainty.
This suggests that they could be detected and automatically corrected (by setting to zero the corresponding values in the density map and disregarding their contribution to the people count), thus improving the accuracy of both the density map and the people count.
As an example, a simple policy for automatic rejection of such regions may consist in rejecting isolated regions with high pixel-level uncertainty/prediction ratio, but other solutions are conceivable. Regardless of the policy, it is intriguing to see that such automatic rejection mechanisms are theoretically possible and practically applicable by integrating dropout injection in the toolbox of video surveillance software.
A further investigation of this issue is an interesting topic for future work.

\section{Conclusions}
\label{sec:conclusions}

This work provides the first investigation of injected dropout for epistemic uncertainty evaluation in order to assess if and how this method can be a practical alternative to the current use of embedded dropout which needs time-consuming uncertainty-aware training (i.e., should be designed at training time).
Our analysis focused on regression problems and encompassed two different dimensions related to the behavior of injected dropout, namely prediction accuracy, measured in terms of RMSE, and quality of the uncertainty measure, evaluated in terms of NLL and MA (i.e., calibration).

We analyzed, theoretically and empirically, how prediction error and quality of the uncertainty measure behave as a function of the dropout rate and proposed a novel injection method grounded on the results of our analysis.

Experimental results on several benchmark data sets, including a challenging computer vision task, provided clear evidence that dropout injection can be an effective alternative to embedded dropout, also considering that it does not require a computationally costly re-training procedure of already-trained networks.
Our results also confirm that the effectiveness of injected dropout critically depends on a suitable re-scaling of its uncertainty measure.

We point out two limitations of our analysis, that can be addressed in future work:
(i) We considered only regression problems; an obvious extension is to analyze dropout injection for classification problems.
(ii) We considered the same dropout rate across different network layers; however, our analysis can be easily extended to the case of different dropout rates across the layers, which is known to be potentially beneficial to Monte Carlo Dropout~\cite{Gal2017concretedrop}.

Finally, another interesting issue for future work is to investigate the effectiveness of injected dropout for detecting out-of-distribution samples, which is potentially useful under domain shifts.

\section*{Acknowledgments}
\label{sec:acknowledgments}

This work was partially supported by the projects: ``Law Enforcement agencies human factor methods and Toolkit for the Security and protection of CROWDs in mass gatherings'' (LETSCROWD), EU Horizon 2020 programme, grant agreement No. 740466; ``IMaging MAnagement Guidelines and Informatics Network for law enforcement Agencies'' (IMMAGINA), European Space Agency, ARTES Integrated Applications Promotion Programme, contract No. 4000133110/20/NL/AF;
``Science and engineering Of Security of Artificial Intelligence'' (S.O.S. AI) included in the Spoke 3 - Attacks and Defences of the Research and Innovation Program PE00000014,  “SEcurity and RIghts in the CyberSpace (SERICS)”, under the National Recovery and Resilience Plan, Mission 4 "Education and Research" - Component 2 "From Research to Enterprise" - Investment 1.3, funded by the European Union - NextGenerationEU.

Emanuele Ledda is affiliated with the Italian National PhD in Artificial Intelligence, Sapienza University of Rome. He also acknowledges the cooperation with and the support from the Pattern Recognition and Applications Lab of the University of Cagliari.



\bibliographystyle{elsarticle-num} 
\bibliography{cas-refs}


\appendix
\section{Proof of Eq.~\eqref{eqn:ideal_uncertainty}}
\label{appendix:A}
In the following we prove that, given a data set $\mathcal{D} = \{(x_n, y_n)\}_{n=1}^N$ for a regression problem, and denoting the corresponding predictions of a given regressor by $\hat{y}_n$, the ideal uncertainty measure $\sigma^{*2}$, i.e., the one that minimizes the NLL, is obtained when the following condition holds:
\begin{equation}
    \sigma^{*2}_n = (y_n-\hat{y}_n)^2, \quad n = 1, \ldots, n \ .
\end{equation}

Consider any instance $x_n \in \mathcal{D}$.
The derivative of the NLL with respect to the corresponding $\sigma^2_n$ is:

\begin{eqnarray}
    \frac{\partial \left(\frac{1}{2} \cdot\frac{(y_n-\hat{y}_n)^2}{\sigma^2_n}+\frac{1}{2} \cdot \log(\sigma^2_n)\right)}  {\partial \sigma^2_n} = \\
    \frac{1}{2} \cdot
    \left(
    (y_n-\hat{y_n})^2 \cdot \frac{\partial (\frac{1}{\sigma^2_n})}{\partial \sigma^2_n} + 
    \frac{\partial (\log(\sigma^2_n))}  {\partial \sigma^2_n}
    \right) = \\
    \frac{1}{2} \cdot
    \left(
    \frac{-(y_n-\hat{y}_n)^2}{(\sigma^2_n)^2} + 
    \frac{1}{\sigma^2_n}
    \right) = \\
    \frac{\sigma^2_n-(y_n-\hat{y}_n)^2}{2(\sigma^2_n)^2}
\end{eqnarray}

Setting the above derivative to $0$ and solving for $\sigma^2_n$ we obtain:
\begin{equation}
    \frac{\sigma^2_n-(y_n-\hat{y}_n)^2}{2(\sigma^2_n)^2} = 0 \ ,
\end{equation}

\begin{equation}
    \sigma^2_n = (y_n-\hat{y}_n)^2, \quad \sigma^2_n \neq 0
\end{equation}





\end{document}